# Weighted Bayesian Gaussian Mixture Model for Roadside LiDAR Object Detection

Tianya Zhang, *Ph.D.*, Yi Ge, Peter J. Jin, *Ph.D.*

*Abstract—* Background modeling is widely used for intelligent surveillance systems to detect moving targets by subtracting static background components. Most roadside LiDAR object detection methods filter out foreground points by comparing new data points to pre-trained background references based on descriptive statistics over many frames (e.g., voxel density, number of neighbors, maximum distance). However, these solutions are inefficient under heavy traffic, and parameter values are hard to transfer from one scenario to another. In early studies, the probabilistic background modeling methods commonly used for the video-based system were considered unsuitable for roadside LiDAR surveillance systems due to the sparse and unstructured point cloud data. In this paper, the raw LiDAR data were transformed into a structured format based on the elevation and azimuth value of each LiDAR point. With this tensor representation, we break the barrier to allow the efficient multivariate Gaussian Mixture Model (GMM) for LiDAR background modeling. The Bayesian Nonparametric (BNP) approach integrates the intensity value and 3D measurements to exploit the measurement data using 3D and intensity info entirely. An adaptive GMM was also implemented that can process LiDAR background modeling in real-time. The proposed method was compared against two state-of-the-art roadside LiDAR background models, computer vision benchmark, and deep learning baselines, evaluated at point, object, and path levels under heavy traffic and challenging weather. This multimodal Bayesian GMM can handle dynamic backgrounds with noisy measurements and substantially enhances the efficiency of infrastructure-based LiDAR object detection, whereby various 3D modeling for smart city applications could be developed.

## I. Introduction

LiDAR (Light Detection and Ranging) has promising potential to benefit future roadside sensor networks. For intersection analytics, wherein most crashes happen, one LiDAR device could cover an entire intersection area with accurate 3D distance measurement. While it usually needs to have multiple cameras/radars to cover one intersection from different angles. The LiDAR solution does not incur privacy concerns; therefore, it is suitable for privacy-sensitive applications, such as crowd management and vehicle tracking. LiDAR sensors can function in various environments, unlike the camera sensor, which relies on external illumination. The data collected through LiDAR could be used to optimize signal control and reduce delay and emission, as well as people counting for event planning. Combined with the Vehicle to Infrastructure (V2I) communication, the 3D data could be fed into many connected vehicle applications, such as Red-Light Violation Warnings and Queue Warnings (Q-WARN), to reduce congestion and prevent probable collisions. Though LiDAR is currently a high-maintenance device, as a critical perception component of autonomous vehicles, the LiDAR technology evolves rapidly. Thanks to the industrial and academic communities enthusiasm for self-driving technology, it will soon become a worthy investment.

Difficulties arise, however, when applying roadside LiDAR sensors for collecting high-resolution micro traffic data. One of the biggest challenges is the large volume of 3D data that is too sparse and disordered to process. This leads to the non-negligible cost of data transmission and processing. Each LiDAR unit produces millions of data points per second, a significant barrier to raw data handling and computing. For roadside applications, the best approach for reducing the data workloads is to separate the backgrounds and foregrounds and only keep the targeted objects. Existing roadside LiDAR methods are reference-based, which store an array of background references (e.g., a list of voxels, point clouds, or statistic descriptors) after eliminating moving objects from aggregated frames. The learned background statistics detect moving objects by comparing new data frames with the background thresholds. These reference-based methods are usually created based on a particular aggregated indicator, such as the number of neighbors, slopes, voxel point density, and maximum range, which is often inappropriate for multimodal backgrounds. From the theoretical perspective, many well-developed dynamic background models for pixel images are undervalued and considered unsuitable for LiDAR point clouds. This research attempts to apply the probabilistic-based background modeling method to the roadside LiDAR application. Unlike the projection-based methods that project 3D point clouds on a range heatmap or the Bird-Eye-View (BEV), e.g., PIXOR [1], our proposed method overcomes the sparsity of LiDAR data by converting raw data into a tensor format given the spherical angular coordinate. In addition, A Bayesian framework is built to fully utilize the reflective intensity and 3D measurements of the point cloud.

This paper successfully implements well-developed multimodal background modeling methods for roadside LiDAR

This work is funded by the New Jersey DOT Real-time signal performance measures (Project No. 2016-14); New Brunswick Innovation Hub Smart Mobility Testing Ground (SMTG) Contract Numbers: 21-60168.

T. Zhang is a Postdoc at Rutgers University, NJ 08854 USA (e-mail: tz140@scarletmail.rutgers.edu).

Yi Ge is PhD candidate at the Civil and Environmental Engineering Department in Rutgers University, Piscataway, NJ 08854 USA (e-mail: yi.ge@rutgers.edu).

P. J. Jin, is associate professor at Civil and Environmental Engineering Department in Rutgers University, Piscataway, NJ 08854 USA (e-mail: peter.j.jin@rutgers.edu).

object detection, transforming 3D point clouds into a structured and compact representation according to elevation and azimuth angular values. A Gaussian Mixture Model (GMM) is created for each elevation-azimuth grid to model the probability of background $P(Background|x)$ given data point $x$. The Bayesian framework is applied to learn the number of mixture components from the data with the Dirichlet Process and weighted samples.

Major contributions are summarized as follows:
1. The method resorts to the high-order tensor data structure, breaking the barrier to allow the flexible and probabilistic approach to analyze LiDAR data.
2. The multivariate GMM for each spherical angular grid makes LiDAR background subtraction as efficient as video-based background subtraction.
3. We applied the nonparametric Dirichlet Process to estimate the number of mixture components for each elevation-azimuth grid so that it can overcome the over- or under-fitting issues.
4. A Weighted Bayesian framework is proposed by fusing the intensity and 3D measurement of LiDAR data to increase the robustness.

## II. RELATED WORK

### A. LiDAR Object Detection

#### 1) Supervised Learning Methods

Recent years have already seen many effective deep learning-based LiDAR object detection models developed for self-driving vehicles. Depending on the data representation methods, the deep learning models can be classified into the Voxel-based methods [2, 3], Point-based methods [4], Frustum-based methods [5], Pillar-based methods [6], and 2D projection-based methods [7~9]. Another way to classify the LiDAR object detection model is whether it uses the two-stage region proposal network or a one-stage framework [10]. For example, the Yolov3d model builds a 3D object bounding box detection method on the Yolov2 image-based one-shot regression meta-architecture [11]. PointRCNN [12] is a typical region proposal network that generates a small number of high-quality 3D proposals to segment foreground objects and then refines the 3D proposal in the second state to obtain final detection results. A weekly supervised framework [13] for 3D point cloud object detection and annotation is developed to address the costly labeling process of supervised learning. This weekly unsupervised framework only takes simple clicks to label the object center on BEV and then uses the BEV detection to reconstruct 3D parameters with a cascade network. Zhou et al. [14] developed a practical framework that outputs both 3D Instance Segmentation and Object Detection results with Spatial Embedding by considering both the global Bounding Box and local point information. By Considering temporal information in consecutive frames, Graph Neural Network and Spatiotemporal Transformer Attention are applied for 3D object detection from point clouds [15]. Chen et al. [16] used range images generated from current and past 3D LiDAR scans as inputs and outputs for labels indicating moving objects in the current range image. Li et al. [17] proposed a one-stage anchor-free 3D LiDAR vehicle detection framework consisting of a voxel/pillar feature extractor, backbone encoding layers, and detection head layers. In a paper [18], a modified convolutional architecture adds dense connections to the convolutional layers for better feature extraction. This allows the deep learning model for roadside LiDAR applications after training on an autonomous driving dataset.

#### 2) Unsupervised Learning Methods

Given the stationary scenario of roadside LiDAR, most existing roadside LiDAR object detections are through background-filtering-based moving object detection [19]. A review paper [20] summaried roadside LiDAR characteristics, single and multiple LiDAR perception methods, and existing datasets to cover the current issues and trends of usage of roadside LiDAR. The 3D-density-statistic-filtering (3D-DSF) model [21] considers the points density of LiDAR cloud points in the voxelized 3D space and applied frame aggregation, point density statistics, and threshold learning at different ranges to find the background cubes. Image-based methods project LiDAR onto the X-o-Y plane and use the image-based background model to detect moving objects at the cost of missing the Z-value of 3D measurements [22]. A variable dimension-based method [23] was developed to store background points instead of storing background voxels by utilizing the static properties of neighbor points. The searching distance is used to judge whether a point belongs to the neighbor of the particular point data. Roadside Lidar was also used for detection and classification at intersections by analyzing the position, velocity, and direction of pedestrians and vehicles [24]. The main drawbacks of voxel-based methods stem from the difficulty of selecting the voxel size that influences the accuracy and computational load. A background filter method is developed to classify truck body types by analyzing point density for each horizontal cell [25]. Based on the operational principle of the LiDAR device, discrete horizontal and vertical angular values were considered as coordinates of pixels in the image matrix. The mean and maximum distance for each azimuth were used to construct background references by analyzing maximum and mean distances over accumulated frames with clean backgrounds [26]. This paper [26] was similar to our approach and selected as one of the baseline models. In the reference [27], the authors developed a height azimuth background construction method considering the height changes in the presence of moving targets for each azimuth grid. A vehicle speed estimation method [28] is explored through background removal, moving point clustering, and vehicle classification, which is based on the max-distance, assuming that the static environment is the furthest point of each laser beam. An unsupervised point cloud pre-training framework ProposalContrast [29], is proposed that learns robust 3D representations by contrasting region proposals. This framework optimizes inter-cluster and inter-proposal separation to utilize the discriminativeness of proposal representations across semantic classes and object instances.

## B. Problems

The deep learning method dominates LiDAR object detection due to its remarkable feature learning capability. However, neural network models are very data-hungry. Deep learning models are usually trained on a curated dataset to generalize across scenes. This brings the issue that they can only reliably function under similar scenarios and often don't work for the unseen dataset. Despite the high potential of the deep learning approach, these models' structures lack transparency. Recently, the roadside LiDAR began to gain momentum as a new measure of traffic data collection for safety analysis and connected vehicle applications [30]. Compared to the self-driving LiDAR models developed for a robot to explore and understand its ever-changing environment, the roadside LiDAR detects moving objects in a fixed setting.

The earlier roadside LiDAR background filtering methods were designed from engineering viewpoints to filter out background voxels/points according to aggregated descriptors. Existing methods are sensitive to parameter settings, such as voxel size, density, and distance thresholds. Understanding the unique characteristics of the LiDAR sensor is necessary before diving into the methodology part. First, the LiDAR data contains non-returnable points when objects exceed the range limit. LiDAR data has a dual-return mode, meaning only a part of the laser pulse comes across an object and reflects from there, whereas the rest of the pulse keeps traveling till it encounters an object. Second, the stationary background objects in LiDAR data can also tremble due to the rotary pulsed laser beams. Third, the LiDAR sensor emits in sequential order at each spin, and the angular values of the same laser beam between two consecutive frames often drift. The offset is larger than the azimuth resolution. Last, LiDAR data is unstructured, meaning shuffling the 3D points does not change the data. In comparison, the image is structured, which will be different after modifying the order of the pixels. Since most background modeling requires structured input, previous studies could not apply the per-pixel background model to 3D LiDAR data. The existing challenges and problems require a more adaptable and efficient LiDAR background modeling approach to support roadside LiDAR applications.

## III. METHODOLOGY

LiDAR emits multiple laser beams at each spin as a modulated laser system, which can be interpreted based on horizontal and vertical angular spherical coordinates. In this section, we will first reorganize the 3D LiDAR data into structured tensor representation and then fit the reorganized LiDAR data with Bayesian Gaussian Mixture Models for each elevation-azimuth grid. Finally, to fully utilize all measurements from LiDAR data, a weighted sampling approach is devised to fuse intensity and 3D measurements.

### A. Data Transformation

Raw LiDAR data contains 3D measurements, intensity values, and GPS information. For object detection purposes, unstructured LiDAR point clouds need to be transformed into structured representations to enable convolutional or matrix operations. Due to the manufacturing feature, the LiDAR devices don't emit at the fixed angular position for each spin. Therefore, it is not sensible to put LiDAR data points according to their beam ID and sequential firing order. Based on the operational principle of the LiDAR device, each spinning of the LiDAR sensor is considered as a set of spatial slices sampling in the 3D world. Here we apply a hash function to store LiDAR points into stacked spatial slices to transform the point clouds into a structured representation. The functions converting cartesian coordinates into spherical coordinates based on the range ($r$), elevation ($\omega$), and azimuth $\alpha$ angular are described by equation (1) ~ equation (3).

$$x_a = r_a \times \cos(\omega_a) \times \sin(\alpha_a) \quad (1)$$
$$y_a = r_a \times \cos(\omega_a) \times \cos(\alpha_a) \quad (2)$$
$$z_a = r_a \times \sin(\omega_a) \quad (3)$$

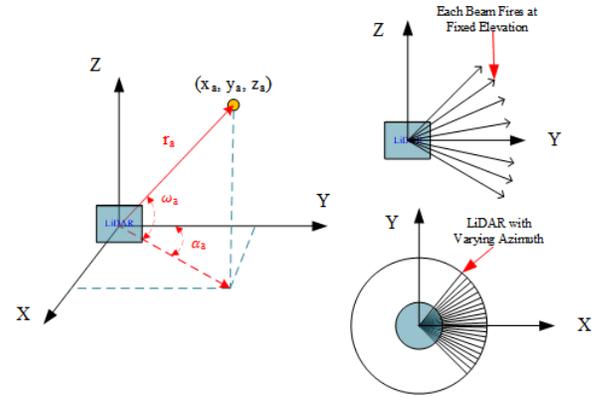

Figure 1 LiDAR Sensor Coordinate System

This study transforms LiDAR data into the tensor structure from the viewpoint of spatial slices at different vertical and horizontal angular. The LiDAR beams are emitted at a fixed vertical angular resolution, and the rotation frequency determines the horizontal angular resolution (See Figure 1). Therefore, we can divide the LiDAR points into each vertical angular and horizontal angular unit. Since the point's vertical angular is associated with fixed beam ID, we only need to identify the proper horizontal angular unit based on rotation frequency. For instance, the horizontal angular resolution for 10 Hz LiDAR is 0.2°, and we will store the data in 360/0.2=1800 different grids. The hash function maps the horizontal angular (α) into an index of the azimuth angular grid.

$$h(\alpha) = \mathrm{mod}\left(\left\lfloor \frac{\alpha}{\text{Azimuth Resolution}} \right\rfloor + 1, 1800\right) \quad (4)$$

Where $\alpha \in (0, 360]$.

Instead of dividing the LiDAR model into voxel cells, this operation will arrange the data with structured spherical angular representation according to the beam ID and azimuth value. Eventually, the LiDAR points will be in two separate data structures; one stores the intensity information, and the other stores 3D measurements (Figure 2).

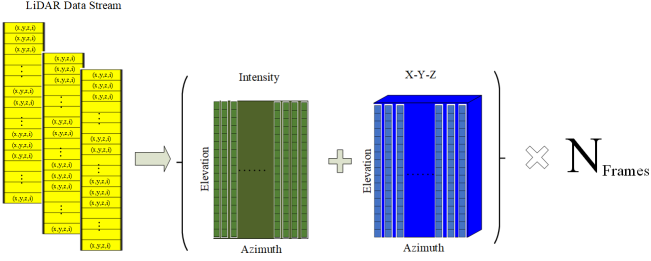

Figure 2 Lidar Data Rearrangement

Processing the LiDAR data in spherical coordinates is consistent with its functioning mechanism. This transformation stored LiDAR points in intensity and 3D measurement tensors and is expandable to include range information. Since the elevation is a fixed value for each beam, it is considered a known parameter. The intensity tensor has the size of $Azimuth\ Grids * Beams * N$; For the $X - Y - Z$ tensor, the size is $Aimuth\ Grids * Beams * 3 * N$, where $N$ is the total frame number. We can draw an analogy between R-G-B channels and grayscale in computer vision with LiDAR's X-Y-Z and intensity value. At this point, the unstructured LiDAR point clouds are transformed into a structured format.

*B. Bayesian Gaussian Mixture Model*

The GMM models [31 ~ 34] have been developed for decades and are implemented with many skills and experience for satisfactory real-time performance. Fortunately, compared to the camera sensor, the LiDAR sensor does not experience moving shadows, camouflage (foregrounds have similar colors to the background), sudden illumination changes, and challenges like that, which is an excellent method for handling dynamic environments. The probabilistic GMM model with K components that learns the background subcomponents is described as follows.

$$p(x|\theta_1, \ldots, \theta_K) = \sum_{j=1}^{K} \pi_j N(x|\mu_j, S_j) \quad (5)$$

The index $i = 1, \ldots, n$ represents observations; index $j = 1, \ldots, k$ represents components. $\theta_j = \{\boldsymbol{\mu}_j, S_j, \boldsymbol{\pi}_j\}$ is the parameter for $j^{th}$ component. $\boldsymbol{\pi}_j$ is the mixing proportion for component $j$. $\boldsymbol{\mu}_j$ is $j^{th}$ mean vector and $S_j$ is the inverse covariance matrix.

Traditional parametric models often suffer from over or under-fitting data due to a fixed and finite number of parameters. The Bayesian Nonparametric (BNP) Modeling is an alternative approach to mitigate underfitting/overfitting with unbounded complexity and full posterior approximation. Due to its richness, computational ease, and interpretability, Dirichlet Process (DP) is one of the most popular methods for BNP models across statistics and machine learning. The Dirichlet process mixture models are also known as infinite mixture models, a distribution over distributions.

The Dirichlet Process is expressed as follows:

$$\begin{aligned} x_i | c_i, \Theta &\sim N(\mu_{c_i}, S_{c_i}) & (6) \\ c_i | \pi &\sim Discrete(\pi_1, \ldots, \pi_K) & (7) \\ (\mu_j, S_j) &\sim G_0 & (8) \\ \pi | \alpha &\sim Dir(\frac{\alpha}{K}, \ldots, \frac{\alpha}{K}) & (9) \end{aligned}$$

Where $G_0$ is the joint prior distribution, $\alpha$ is a positive scaling parameter. $c_i$ ($i = 1, \ldots, n$) is a stochastic indicator variable to encode the component to which observation $y_i$ belongs. The distribution of the number of observations assigned to each component is named occupation number, which is multinomial given mixing proportions $\pi$:

$$p(n_1, \ldots, n_K|\pi) = \frac{n!}{n_1! n_2! \ldots n_K!} \prod_{j=1}^{K} \pi_j^{n_j} \quad (10)$$

The distribution of indicators is expressed as follows:

$$p(c_1, \ldots, c_n|\pi) = \prod_{j=1}^{K} \pi_j^{n_j} \quad (11)$$

The construction of the Dirichlet Process is called the Stick-Breaking method. Suppose there is a unit-length stick. Let $\beta_c \sim Beta(1, \alpha)$ for $c = 1, 2, 3, \ldots$, and regard $\beta_c$ as fractions for how much we take away from the remainder of the stick every time. The smaller $\alpha$ is, the fewer the stick will be left (on average) and leading to more concentrated distribution. Then $\pi_c$ can be calculated by the portion of the remaining stick we take away each time.

$$\pi_c = \beta_c \prod_{j=1}^{c-1} (1 - \beta_j) \quad (12)$$

The implementation of stick-breaking is through the Chinese Restaurant Process (CRP). The Chinese restaurant metaphor is used to represent each sample as a customer who arrives and takes a seat at a table. The tables ($T$) represent Gaussian components, and a customer can choose to sit at a table ($t$) where others are already seated with a probability that is proportional to the number of customers already there. Dishes on the menu represent the distribution of values (base measures) for each table (mixture component). Alternatively, the customer can choose to sit at a new table with a probability that is proportional to $\alpha$. By integrating the draw from the DP, the infinite set of sticks in the Chinese restaurant process is replaced by a finite set of tables that can be computed more efficiently.

The probability of a given data point belonging to the background ($B$) is through marginalizing all mixture components $T$:

$$P(x|B) = \sum_{t \in T} P(x|t, B) \quad (13)$$

$$P(x|t, B) = \frac{s_t}{\sum_{i \in T} s_i} P(x|n_t, k_t, \mu_t, \sigma_t^2) \quad (14)$$

Where $t$ is the $t^{th}$ Gaussian component (table). $n_t, k_t, \mu_t, \sigma_t^2$ are student-t parameters for each component $t$. The total number of $s_t$ samples are clustered as $t$.

Then we can apply the Bayesian rule for background (B) /foreground (F) classification by computing the probability of sample $x$:

$$P(B|x) = \frac{P(x|B)P(B)}{P(x|B) + P(x|F)} \quad (15)$$

Based on the DPGMM (Dirichlet Process Gaussian Mixture Model) implementation in the reference [35], $P(x|F) = 1$ is considered a uniform distribution. $P(x|B)$ is obtained from equation (11), and $P(B)$ is the threshold value.

## C. Weighted Dirichlet Process Gaussian Mixture Model

LiDAR sensors generate precise 3D measurements without distortion issues compared to camera devices. Because LiDAR intensities are not as discriminative as cameras, most LiDAR-background filtering methods use distance info, and the intensity value has often been overlooked. Some studies [36, 37] have found that leveraging intensity as an additional feature allows for more accurate object detection. The intensity value of LiDAR is mainly impacted by the reflectivity of the obstacle surface, which is often used for infrastructure detection, e.g., lane markings, traffic signs, and buildings. For roadside LiDAR applications, the intensity values of static infrastructures are relatively stable, which are suitable as contextual information.

In this study, we tackle the problem by fusing the intensity and 3D measurements based on Weighted Dirichlet Process. An extended framework is developed that uses intensity value to generate a prior weight for each data point as backgrounds.

We fit the intensity data per elevation-azimuth grid with a Gaussian Mixture Model (GMM). The number of components is set as $K = 5, 7 \text{ or } 9$, according to the convention by video-based GMM. The induced weight ($\omega_d$) of point cloud data is expressed as:

$$\omega_d = 1 + \text{SamplingRate} * \sum_{k=1}^{K} \delta_k * \omega_k \quad (16)$$

$$\delta_k = \begin{cases} 0, & if\ C_d \neq k \\ 1, & if\ C_d = k \end{cases} \quad (17)$$

Where K is the total number of Gaussian Mixtures. $\omega_k$ is the weight for the $k^{th}$ Gaussian Components. $C_d$ denotes the component classification for point cloud data $d$. The $SamplingRate$ can be set as $\{0, 2, 4, 8\}$ so that it is backward compatible to unweighted data. The larger the $SamplingRate$ value, the faster the Dirichlet Process will converge. This approach can be considered as a robust Bayesian nonparametric method, which down-weights outliers using intensity as the prior. The Weighted DPGMM is a flexible extension of DPGMM through a weighted sample design.

## D. The Roadside LiDAR Detection Framework

The workflow is broken down into steps (see Figure 3). The first step is to transform LiDAR packet data into a tensor format using the hash function. The second step is to determine the detection zone. A large amount of the backgrounds are distant buildings. This can be done with Geofencing to filter out non-drivable or walkable space based on GPS location. Next, the proposed background modeling will be applied to create foreground masks using the combination of intensity and X-Y-Z measurements. The local Outlier Factor (LOF) algorithm is used to remove noisy points that are left out by background modeling steps. After finishing that, a density clustering-based bounding box estimator was used to obtain detections from the segmented objects. Each detected object was encoded into a state space with 3-dimensional coordinates, angles, and speeds. The detected object is classified into different road user groups, including pedestrians, passenger cars, trucks, and large freight vehicles. The object classification is based on detection bounding box 3D measurements, length/height ratio, and traveling speeds. Then a joint probabilistic data association (JPDA) and the interacting multiple model (IMM) filters are applied to track each object over continuous frames [38].

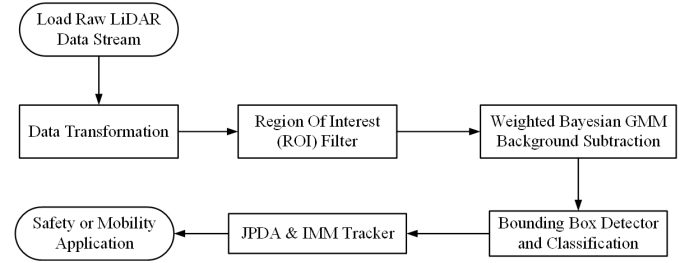

Figure 3 Real-Time LiDAR Object Detection, Classification, and Tracking with GMM Model

## IV. EXPERIMENTAL SETUP

### A. Testing Sites

Two testing sites are selected. Both are installed with 2K cameras and Alpha Prime LiDAR Sensors with communication and power cables. The first site is at the intersection between Albany Street and George Street in downtown New Brunswick, with affinity to many restaurants, banks, and a university campus; The second site is at French street and Joyce Kilmer Ave, which is close to the Robert Wood Johnson hospital with the largest number of employees in this area. The construction plan is a part of the new 2.4-mile DataCity Smart Mobility Testing Ground by Rutgers University's Center for Advanced Infrastructure and Transportation (CAIT) to establish a living laboratory for connected and automated vehicle technologies in New Jersey. Eventually, multiple intersections from the testing corridor will be equipped with high-resolution roadside sensors and edge computing devices to enable innovative mobility applications. The selected intersection layouts and sensors' locations are shown in Figure 4.

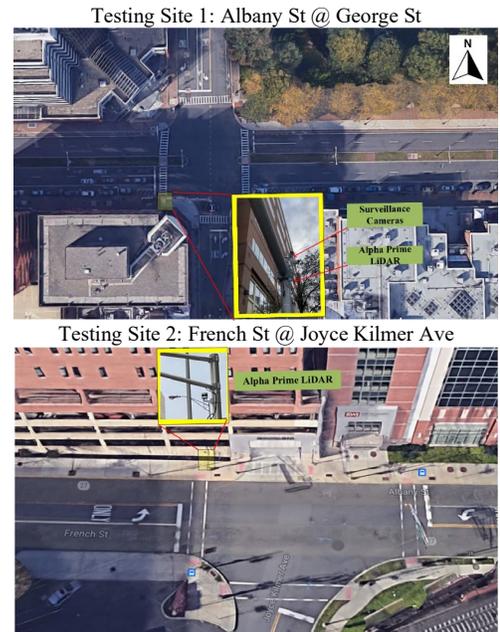

Figure 4 Testing Sites at New Brunswick Smart Mobility Testing Ground

The first testing site is from Albany Street @ George Street, representing a busy urban arterial with traffic signal. The data is collected during morning peak hours on Dec 15, 2021, under heavy traffic conditions. The data from the second testing site, French St @ Joyce Kilmer Ave, were collected on Jan 20, 2022, a snowy day, to investigate how the weather could impact LiDAR performance. We use the 2.7k cameras mounted on the same traffic pole to generate directional counts and compare them against our LiDAR background subtraction methods to evaluate the performance quantitively.

### B. Benchmark Computer Vision Algorithms

The Computer Vision vehicle detection and tracking model is built on the instance segmentation network Mask R-CNN [39] with the backbone of the ResNeXt-101 feature pyramid network [40]. Five object classes of interest, including car, truck, person, bus, and bike. A Weighted Inter-class Non-maximum Suppression (WINS) is applied to remove false interclass detections, as small trucks could result in car and truck double detections. The online association algorithm is employed to assign detections from new frames to existing tracklets, considering both feature similarity and spatial constraints. The outputs of computer vision algorithms are used to evaluate the LiDAR detection results at the trajectory level. The tracked trajectory points from the high-accuracy computer vision algorithms are shown in Figure 5.

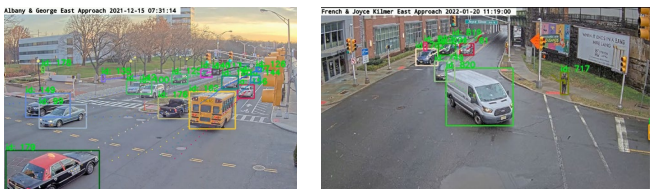

Figure 5 Computer Vision Benchmark Detection and Tracking

## V. RESULTS ANALYSIS

In this section, we will comprehensively analyze the proposed methodology's efficacy from the point, object, and path levels.

### A. Point Level Evaluation

The point-level results are the fundamental step for foreground segmentation. This evaluation part aims to quantify the accuracy of the proposed background subtraction for each LiDAR point. We manually removed background objects from selected LiDAR data frames using the *CloudCompare* software to get ground truth data. The LiDAR frames are chosen randomly at the 200-frame interval to ensure fairness. Each point from LiDAR detection results and ground truth data can be classified as a true positive, true negative, false positive, or false negative. Four performance metrics were used, including Accuracy, Precision, Recall, and F1 score, which were defined in the following equations. Two state-of-the-art roadside LiDAR background modeling methods were implemented against the same ground-truth dataset as baseline models.

$$Accuracy = \frac{TP+TN}{TP+TN+FP+FN} \qquad (18)$$

$$Precision = \frac{TP}{TP+FP} \qquad (19)$$

$$Recall = \frac{TP}{TP+FN} \qquad (20)$$

$$F1\ Score = 2 * \frac{Precision * Recall}{Recall + Precision} \qquad (21)$$

The two baseline models are chosen because they are like our method built on spherical coordinates. The first baseline model, Coarse-Fine Triangle Algorithm (CFTA), applies histogram triangle thresholding to automatically find the dividing value between foreground targets and background objects. The second baseline model is a two-step approach to establish the background range for each elevation-azimuth value using the mean and max values calculated from many frames. As shown in the TABLE I, our model outperformed the baseline methods by all metrics in two scenarios. First, all models' accuracy is very high because the background and foreground point clouds are highly imbalanced. According to our 128-beam LiDAR dataset, the background objects account for 95% of the total point clouds (Figure 6). The high accuracy scores suggest that all methods can remove most background components correctly.

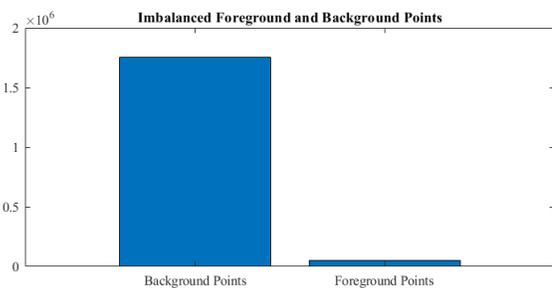

Figure 6 Imbalanced Backgrounds and Foregrounds in 128 Beam Roadside LiDAR Point Clouds

Due to the impact of snow, the foreground detection results on Site-2 show lower Precision than on Site-1. The snowstorm exerts the most significant influence on the Mean-Max model that uses basic mean and maximum values for background construction. While the Bayesian GMM method demonstrates better robustness under snowy weather among all three methods, evidenced by the highest Precision score on the Site-2 intersection. Overall, our proposed algorithm has the best performance in all four categories (See TABLE I).

TABLE I
POINT LEVEL EVALUATION RESULTS

|  | Testing Site 1: George @ Albany | | | | Testing Site 2: French @ Joyce Kilmer | | | |
|---|---|---|---|---|---|---|---|---|
|  | Accuracy | Precision | Recall | F1score | Accuracy | Precision | Recall | F1score |
| DPGMM | **98.8%** | **96.1%** | **93.6%** | **94.8%** | **99.5%** | **84.9%** | **95.2%** | **89.8%** |
| CFTA | 98.0% | 90.2% | 92.3% | 91.2% | 98.6% | 64.1% | 79.9% | 71.1% |
| MeanMax | 90.9% | 55.0% | 85.1% | 66.8% | 94.2% | 22.8% | 72.9% | 34.7% |

In Figure 7, the foreground segmentation results are presented. The scatted points in the air are falling snow. The phantom LiDAR points inevitably impact all three models under challenging weather. The proposed model is the most efficient among all comparable methods, with the cleanest background points.

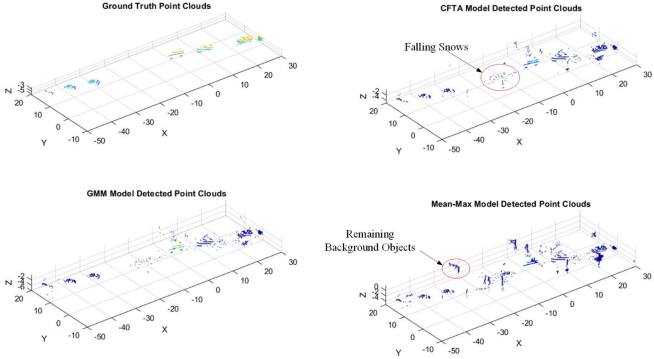

Figure 7 Point Level Foreground Segmentation Evaluation

### B. Object Level Evaluation

We conducted the 3D bounding box detection assessment for object-level assessment from randomly selected data frames at 300 intervals. We reviewed the extracted frames with 3D object bounding box detections and compared the bounding box outputs to the original LiDAR recording to judge if a bounding box was correct or if any missed detection happened. Based on the segmented foreground points, the object-level evaluation evaluates how the occlusion and bad weather could impact the object detector. The validation of object-level detection is summarized with the number of True Positives, False Negatives, and False Positives, which are then used for calculating Precision, Recall, and F1 Score (See TABLE II).

For Site-1, the occlusion issues are significant during the red signal phase, resulting in insufficient LiDAR points for the detector to create a bounding box. Therefore, the Recall score at Site-1 is low. From all the selected frames, very few false positives were found in Site-1, showing the proposed model can efficiently remove background points. For Site-2, the vehicle volume is low, causing fewer occlusions. Due to the impact of the snowstorm, the precision score on Site-2 is lower than the Site-1 intersection. The density-based detector falsely generated bounding boxes for the falling snows when phantom LiDAR points were larger than the detector's threshold.

TABLE II
OBJECT LEVEL EVALUATION RESULTS

|  | Testing Site 1: Albany @ George | Testing Site 2: French @ Joyce Kilmer |
|---|---|---|
| Precision | 99.03% | 97.41% |
| Recall | 78.21% | 90.40% |
| F1 Score | 87.39% | 93.78% |

Figure 8 displays major occlusions observed at the signalized intersection caused by the queuing vehicles during the red-light phase. Vehicles on the remoter lane are often blocked by cars on the closer lane. However, these blocked vehicles could be detected again after the signal turns green.

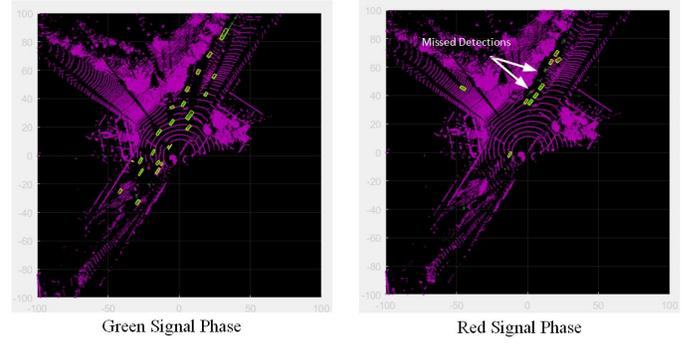

Figure 8 Object Level Detection Results Evaluation

To further demonstrate the advantage of unsupervised learning vs. deep neural networks for roadside LiDAR, we tested several Deep Learning (DL) frameworks [42~44] pre-trained on the public dataset. As shown in Figure 9, although the DL model attains valid results on a self-driving dataset, the pre-trained model cannot be directly used for unseen dataset. Preparing new training dataset from the ground up or transfer learning for a new scenario could be grueling and costly. Given the generalization issue of supervised learning approaches, the Bayesian mixture modeling framework does not need labeled training data and is a sensible choice for infrastructure LiDAR.

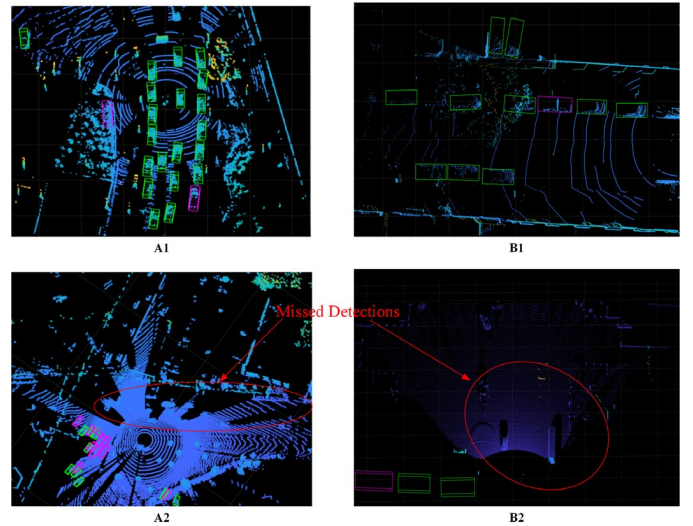

Figure 9 Deep Learning Baselines on a pre-trained dataset and our Roadside Data (A1 PointPillar on PandaSet; A2 PointPillar on Our Roadside LiDAR; B1. Complex-YOLOV4 on PandaSet; B2. Complex-YOLOV4 on DAIR-V2X Roadside LiDAR [41])

### C. Path Level Evaluation

The previous point level and object level evaluations are performed on static LiDAR frames. In contrast, the path level evaluation examines the tracking module performance with bounding box detections for continuous LiDAR frames. Figure 10 presents the segmented foreground using DPGMM and bounding box detection results side-by-side. The red boxes are bounding box measurements in 3D dimensions with an orientation angle. The green box is tracked object based on historical records. In our settings, if the bounding box is observed consecutively for six frames, we will change that object's status from candidates to confirmed objects. The tracked object is missing seven frames in the past eight frames, and then the algorithm will delete the objects from the tracking records. The algorithm can efficiently overcome the occlusion problem when tracked vehicles are temporarily obstructed by cars in the adjacent lane.

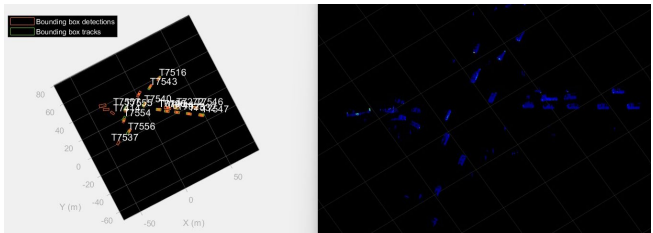

Figure 10 Bounding Box Detection and Tracking on Segmented Moving Vehicles

For the selected intersection, each camera mainly faces one approach. We choose the approach with the highest traffic volume and compare the inbound and outbound traffic volume within 15-min to examine the path-level object detection results. The computer vision model [45] is a winning CVPR AI City Challenge solution for traffic camera movement counts. The parameters were maintained the same in this experiment, except that the original version detects three classes while our implementations have five categories. The camera views from two testing sites are shown in Figure 5. Testing results show that our tracking model is accurate in congested and challenging weather conditions. All movement counting accuracies are above 92% (TABLE Ⅲ). The extracted trajectories from two locations were plotted in Figure 11, containing vehicles and pedestrians. The trajectory data can be used for analyzing pedestrian safety issues to near-miss events at intersections.

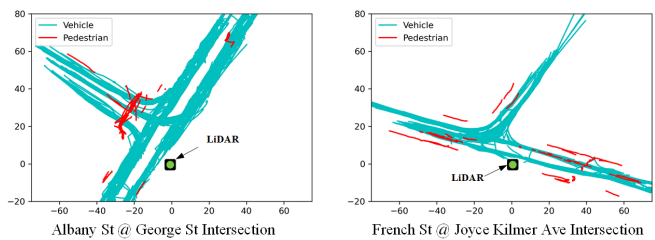

Figure 11 Vehicle and Pedestrian Trajectory Using Proposed Method

TABLE III
PATH LEVEL EVALUATION RESULTS

|  | Testing Site 1: Albany @ George | | Testing Site 2: French @ Joyce Kilmer | |
| --- | --- | --- | --- | --- |
|  | inbound | outbound | inbound | outbound |
| Video Count | 211 | 221 | 86 | 92 |
| LiDAR Count | 223 | 204 | 82 | 88 |
| Accuracy | 94.31% | 92.31% | 95.35% | 95.65% |

## VI. DISCUSSION

As the previous literature [46] discussed, background subtraction using DPGMM can learn a very precise model. Applying the proposed model for roadside LiDAR object detection could largely enhance the efficiency of point cloud processing. For infrastructure LiDAR, the object detection task only pertains to a small amount of data in a fixed environment. Therefore, background modeling can significantly improve the data chain efficiency by only transitioning a tiny portion of the LiDAR point clouds. In our testing experiment, 31.7 GB of raw point cloud data were compressed into 760 MB foregrounds and a background reference, which reduced around 95% redundancy in roadside LiDAR data. Figure 12 uses the proposed method to build the background model as static data and only update the foreground moving vehicle points. That's why it doesn't contain shadows as in the original LiDAR point clouds.

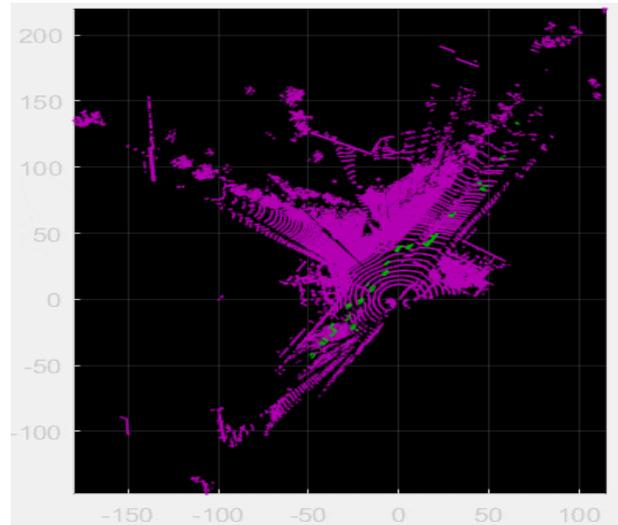

Figure 12 Segmented Moving Object Overlayed on Pretrained Background

However, the Dirichlet Process is computationally inefficient that comes with a time complexity of $O(n)$, where n is the total number of elevation-azimuth grids. For real-time application, it needs massive parallel GPUs. To address the computational constraint, a CPU version using the adaptive GMM model was also implemented, which processes about 15 frames/sec on an 8-Core Ryzen 5800X computer. The real-time adaptive GMM model needs to predefine the number of Gaussian components $K$ and an adaptive learning rate $\lambda$. A small learning rate updates the background slowly; a larger learning rate is prone to misclassify sleeping objects, bringing the missed detection issue when vehicles wait before the red light. The real-time version demonstration can also be found on the repository [47].

This research's outcome will help integrate the infrastructure-based LiDAR into connected intersection applications. Figure 13 plots the Time-Space diagram with high-resolution (0.1s) vehicle trajectory for the westbound approach from the Albany & George intersection to show the cycle-by-cycle traffic flow patterns. We defined the entry location of westbound lanes as the reference site and calculated the relative distance of vehicle position at each timestamp. The time-space diagram is an essential analytic tool for understanding the traffic flow, especially useful for shockwave discussion, queuing formation, and dissipation.

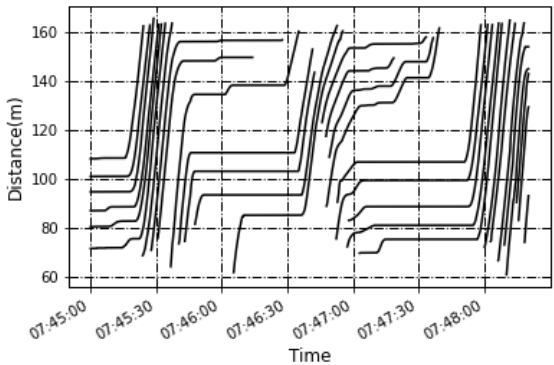

Figure 13 Cycle-by-Cycle Time-Space Diagram from Roadside LiDAR Vehicle Trajectory

Since the LiDAR sensor is installed at a fixed location, we applied point cloud registration to build the coordinate transformation between LiDAR's measurement with the world GPS coordinates in latitude and longitudinal. Figure 14 shows the projected trajectory with GPS locations showing adequate coverage with a single LiDAR sensor over the intersection area.

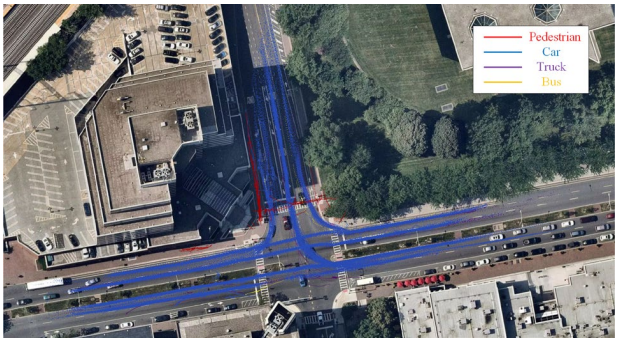

Figure 14 GPS Trajectory using LiDAR Point Registration

## VII. Conclusion

The multivariate probablistic methods have encompassed many classic algorithms throughout machine learning to model data distribution. However, these mature techniques haven't been effectively utilized for roadside LiDAR background modeling. The main reason is that LiDAR data is inherently unstructured and sparse in space, whereas the images are relatively formatted. This research successfully adopted the probabilistic Bayesian GMM model for 3D object detection and was validated at point, object, and path detection levels. Most LiDAR modeling methods are built point-wise or voxel-wise in cartesian coordinates, while this solution rearranges LiDAR data points via spherical coordinates. Due to the sparsity of point cloud in 3D dimension, the voxel-based methods incur unnecessary computational burden and are sensitive to the voxel sizes. In addition, 2D plane projection increases the processing speed at the cost of losing information. By converting raw LiDAR point clouds into the tensor format based on spherical angular coordinates, we can apply background modeling approaches to each LiDAR point just as an image pixel.

Specifically, the weighted sampling is proposed to fuse the intensity and distance measurement. And the multivariate GMM is used at every elevation-azimuth unit to construct the density function of X-Y-Z measurement for background representation. In addition, Dirichlet Process is implemented to automatically determine the number of Gaussian Mixture components. Finally, the conventional adaptive GMM methods were also implemented to process the data in real-time and do not require multi-processing supported by GPU parallelism capability.

This method allows for stable and rapid adaptation to a dynamic environment for an assortment of surveillance applications. The algorithm of the roadside LiDAR model can be readily generalized to different scenarios and empower V2X driving technology and help to build cyber-physical systems digital-twin modeling of smart city.